\pdfoutput=1

\documentclass[11pt]{article}

\usepackage[]{acl}

\usepackage{times}
\usepackage{latexsym}

\usepackage[T1]{fontenc}

\usepackage[utf8]{inputenc}

\usepackage{microtype}

\usepackage{tikz-dependency}
\usepackage{colortbl}
\usepackage{amssymb}
\usepackage{makecell}

\graphicspath{ {./charts/} }

\title{LyS\_ACoruña at SemEval-2022 Task 10: Repurposing Off-the-Shelf Tools for Sentiment Analysis as Semantic Dependency Parsing}

  \author{
  Iago Alonso-Alonso, David Vilares and Carlos G\'{o}mez-Rodr\'{i}guez \\
  Universidade da Coru\~{n}a, CITIC \\
  Departamento de Ciencias de la Computación y Tecnologías de la Información \\
  Campus de Elvi\~{n}a s/n, 15071 \\ A Coru\~{n}a, Spain \\
  {\tt \{iago.alonso,david.vilares,carlos.gomez\}@udc.es} 
  \\}

\begin{document}

\maketitle

\begin{abstract}

This paper addressed the problem of structured sentiment analysis using a bi-affine semantic dependency parser, large pre-trained language models, and publicly available translation models. For the monolingual setup, we considered: (i) training on a single treebank, and (ii) relaxing the setup by training on treebanks coming from different languages that can be adequately processed by cross-lingual language models. For the zero-shot setup and a given target treebank, we relied on: (i) a word-level translation of available treebanks in other languages to get noisy, unlikely-grammatical, but annotated data (we release as much of it as licenses allow), and (ii) merging those translated treebanks to obtain training data. In the post-evaluation phase, we also trained cross-lingual models that simply merged all the English treebanks and did not use word-level translations, and yet obtained better results. According to the official results, we ranked 8th and 9th in the monolingual and cross-lingual setups.

\end{abstract}

\section{Introduction}

\begin{figure}[thbp!]
    \centering
    \small
    \includegraphics[width=\columnwidth]{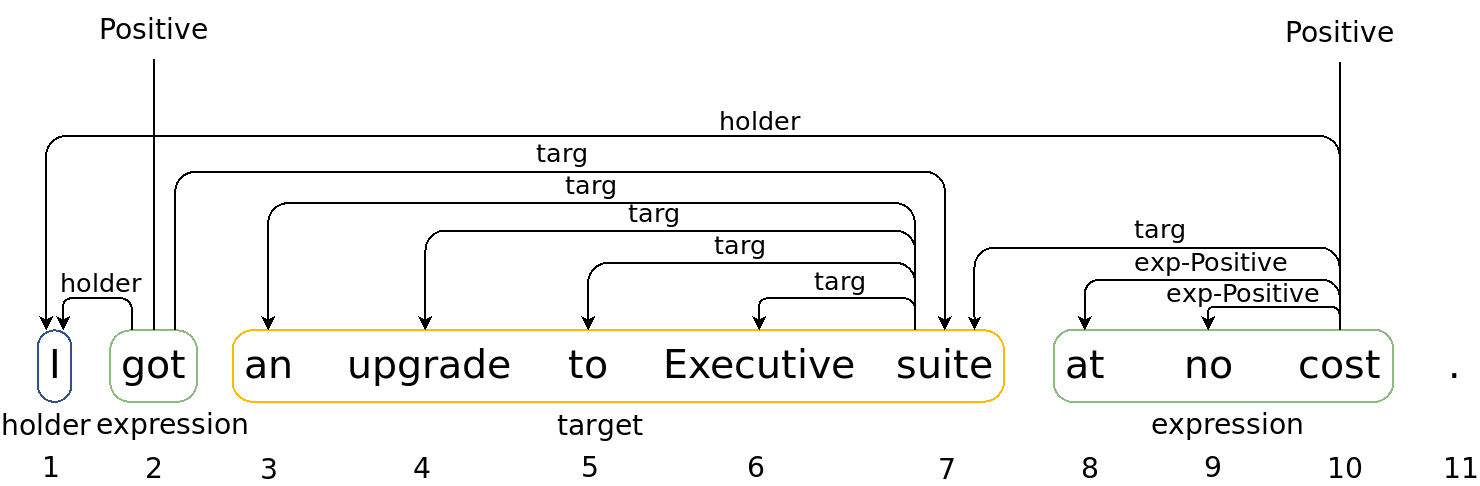}
    \caption{An example of sentiment graph as defined by \citet{barnes-etal-2021-structured}. The sentence has a holder (`I'), two sentiment expressions (`got' and `at no cost') and one target (`an upgrade to Executive suite')}
    \label{fig:sentiment-graph-example} 
\end{figure}

Sentiment Analysis \citep[SA,][]{Pang08opinionmining} deals with the automatic processing of subjective information in natural language texts. Early work on SA focused on conceptually simpler tasks, such as polarity classification at the sentence or document level. With the advances in natural language processing (NLP), more fine-grained and complex tasks have been proposed, such as detecting the entity that expresses an opinionated chunk of text, or the entity that was targeted.
More particularly, \citet{barnes-etal-2021-structured} consider sentiment analysis as a (graph) structured task, and discuss up to five subtasks: (i) sentiment expression extraction, (ii) sentiment target extraction, (iii) sentiment holder extraction, (iv) defining the relationship between these elements, and (v) assigning a polarity label. They discuss that although these tasks have been extensively studied by different authors \citep[\emph{inter alia}]{turney-2002-thumbs,pontiki-etal-2015-semeval,zhang2019end}, they are not addressed all together. They also discuss that such subdivision into subtasks might have a negative impact in the general analysis of the sentence, and that a joint analysis could translate into a holistic approach. To do so, they propose to encapsulate all these tasks in the form of a sentiment graph. Formally, the goal is to find the set of opinion tuples $\{O_1,\ldots,O_i,\ldots,O_n\}$ in a given text, where each opinion $O_i$ is a tuple of the form $(h, t, e, p)$ where $h$ is a holder who expresses a polarity $p$ towards a target $t$ through a sentiment expression $e$, implicitly defining pairwise relationships between elements of the same tuple. We illustrate an example in Figure \ref{fig:sentiment-graph-example}.

More particularly, for the SemEval-2022 Task 10 \cite{barnes-etal-2022-semeval}, the organizers proposed both a monolingual\footnote{We use the term monolingual as it was the term used by the organizers, but this setup allowed the use of any resource, including resources in different languages.} and a cross-lingual (zero-shot) setup. They considered 5 languages (and 7 treebanks): English, Spanish, Catalan, Basque, and Norwegian. For the zero-shot setup Basque, Catalan, and Spanish were the target languages.

\paragraph{Our approach} is based on the idea of viewing this task as semantic dependency parsing \cite{oepen-etal-2015-semeval}, since both tasks are structurally similar even if the graphs have different meaning. More specifically, we rely on a bi-affine graph-based parser \cite{dozat-manning-2018-simpler} and different large pre-trained language models (LM), such as BERT \cite{devlin-etal-2019-bert}, RoBERTa \cite{liu2019roberta} or XLM-R \cite{conneau-etal-2020-unsupervised}. For the monolingual setup we train a semantic parsing model on single and merged treebanks, and compare the performance using different LMs. For the cross-lingual setup, we first do a word-level translation of the datasets in a different language than the target treebank, and then proceed similarly to the monolingual setup. Overall, the approach relies on off-the-shelf tools already available, but traditionally used for other purposes. We here re-purpose them for their use for sentiment analysis as graph-based parsing.

\section{The role of parsing in SA}

Parsing has been used in the past for SA, with different motivations, such as integrating syntactic knowledge as a component of the model's architecture, or producing structured sentiment outputs.

\paragraph{Polarity classification.} Since early times, authors have studied the importance of language structure to deal with relevant linguistic phenomena for polarity classification, first focusing on simpler strategies such as the use of n-grams or lexical rules \cite{pang-etal-2002-thumbs,taboada2011lexicon}. Later on, more complex syntactic structures were incorporated as well, both for rule-based and machine learning approaches. 

For instance, for the rule-based paradigm, \citet{poria2014sentic} used dependency relations for concept-level sentiment analysis, so sentiment could flow from one concept to another to better contextualize polarity. \citet{VilAloGomNLE2015,vilares2017universal} proposed a model to compute the sentiment of sentences that was driven by syntax-based rules to deal with specific relevant phenomena in SA, and that could be easily re-purposed for any language for which a dependency parser was available. \citet{kanayama-iwamoto-2020-universal} built on top of \citeauthor{vilares2017universal}'s idea, and proposed a multilingual syntax-based system that achieved a high precision for 17 languages.

From the machine learning perspective, \citet{joshi2009generalizing,vilares2015usefulness} used dependency triplets to train data-driven (pre-neural) models and obtain slight improvements over purely lexical approaches. \citet{socher-etal-2013-recursive} collected sentiment labels for phrases and sentences that were previously automatically represented as constituent (sub)trees, to then train a compositional model that used a recursive neural network. This work has some relevant resemblances with \citet{barnes-etal-2021-structured}'s proposal for structured sentiment analysis. \citeauthor{socher-etal-2013-recursive} were among the first to provide tree-shaped annotated sentiment data (in this case just for polarity classification), while most of previous work had focused on using tree knowledge as external information to the models, but with sentiment annotations only associated with plain texts. This publicly available data later encouraged many authors to design models that could exploit tree-shaped annotated data to obtain better performing models \cite[\emph{inter alia}]{tai-etal-2015-improved,zhang-zhang-2019-tree}.

\paragraph{Aspect-based sentiment analysis (ABSA).} ABSA is a task that is particularly suitable for the integration of syntactic information, since its main goal is to associate sentiment with specific entities and aspects that occur in the sentence \cite{pontiki-etal-2015-semeval}. Related to this, \citet{popescu-etzioni-2005-extracting} already used dependency trees to constrain an unsupervised sentiment analysis system that extracted a set of product features and their sentiment, given a particular item.
More recently, with the wide adoption of neural networks in NLP, different authors have integrated syntactic knowledge and syntactic structures in different network architectures, such as long short-term memory networks \cite[LSTMs,][]{tang-etal-2016-effective}, recursive neural networks \cite{nguyen-shirai-2015-phrasernn}, convolutional networks \cite{xue-li-2018-aspect}, and graph attention networks  \cite{huang-etal-2020-syntax,sun-etal-2019-aspect}.\\

\noindent Overall, it is clear that parsing has had a high relevance in SA. Yet, the novelty of the shared task is in using graphs to represent richer annotations. This makes it possible to use parsing algorithms as sentiment models, i.e. not just to use them as a component of the model architecture, but as the model responsible of producing the whole sentiment structure of the chunk of text. Also, this is especially relevant in the era of large neural models, where the utility of parsers for downstream tasks is sometimes questioned, with some studies questioning its need in the presence of pretrained models that implicitly learn syntax \cite{tenney-etal-2019-bert,glavas-vulic-2021-supervised,dai-etal-2021-syntax} while others still achieve extra accuracy from their use in conjuntion with such models \cite{sachan-etal-2021-syntax,xu-etal-2021-syntax,li-etal-2021-improving-bert,Zhang2020syntax}. In any case, tasks like this one show that graph structures can be also useful to re-purpose traditional tasks such as SA, while taking advantage of research that the NLP community has done on parsing algorithms for decades.

\section{Brief overview of the shared task}

The goal of the task is to produce graph structures that reflect the sentiment of a sentence, as we showed in Figure \ref{fig:sentiment-graph-example}. More particularly, the organizers released 7 treebanks in 5 different languages: OpeNER \citep[][English and Spanish]{Agerri2013}, MPQA \cite[][English]{Wiebe2005b}, Darmstadt\_unis \citep[][English]{toprak-etal-2010-sentence}, MultiBooked \cite[][Basque and Catalan]{barnes-etal-2018-multibooked}, and NoReC\_fine \cite[][Norwegian]{ovrelid-etal-2020-fine}.\footnote{For more detailed information see \url{https://github.com/jerbarnes/semeval22_structured_sentiment}} Table \ref{tab:detalles-datasets} details the main statistics for the datasets.

\begin{table}[thbp!]
    \centering
    \tiny
    \begin{tabular}{l|ccccc}
        \textbf{Dataset} & \textbf{Language} & \textbf{\# sents} & \textbf{\# holders} & \textbf{\# targets} & \textbf{\# expr.} \\
        \hline
        NoReC\_fine & Norwegian & 11437 & 1128 & 8923 & 11115 \\
        MultiBooked & Basque & 1521 & 296 & 1775 & 2328 \\
        MultiBooked & Catalan & 1678 & 235 & 2336 & 2756 \\
        OpeNER & Spanish & 2057 & 255 & 3980 & 4388 \\
        OpeNER & English & 2494 & 413 & 3850 & 4150 \\
        MPQA & English & 10048 & 2279 & 2452 & 2814 \\
        Darmstadt\_unis & English & 2803 & 86 & 1119 & 1119 \\
        \hline
    \end{tabular}
    \caption{General statistics of the treebanks used in the shared task.}
    \label{tab:detalles-datasets}
\end{table}

\noindent{The sentiment of a sentence is composed of all the opinions, \textit{O\textsubscript{i}}, that make it up. Each opinion can have up to four elements: a holder (\textit{h}) who expresses a polarity (\textit{p}) towards a target (\textit{t}) through a sentiment expression (\textit{e}). These four elements implicitly define the pairwise relationships between the elements of a tuple.} The previous example, Figure \ref{fig:sentiment-graph-example}, shows a sentence with two sentiment expressions (\textit{got} and \textit{at no cost}) that express the polarity (\textit{Positive}) of the sentiment that a holder (\textit{I}) has towards one target (\textit{an upgrade to Executive suite}).

\paragraph{Preprocessing} The organizers of the shared task proposed two possible ways to address the task: as a sequence labeling or as graph-based parsing problem. As mentioned above, we opted for the latter. We use the scripts available in the official repository to transform the JSON files to the CoNLL-U based format and \emph{vice versa}, and we applied the needed changes to make it compatible with \texttt{supar} (see \S \ref{section-model}).\footnote{\url{https://github.com/MinionAttack/conllu-conll-tool}}
Under the graph-based paradigm, the problem is approached as a bilexical dependency graph prediction task, with some assumptions. To convert the data, the organizers suggest two possible conversions, namely \textit{head-first} and \textit{head-final}. In \textit{head-first}, it is assumed that the first token of the sentiment expression is a root node, and that the first token of each holder or target spans is the head node of such span, while the other ones are dependents. Meanwhile, in \textit{head-final}, the final token of the holder and target spans is set as the head of the span, and the final token of the sentiment expression as a root node (Figure \ref{fig:sentiment-graph-example} is a head-final example). In this work, we have chosen \textit{head-final}, which is the default option for the shared task and also delivered better results than \textit{head-first} in the experiments carried out by \citet{barnes-etal-2021-structured} (see Table 3 in that paper).

\paragraph{Subtasks} More in detail, the challenge is divided into two subtasks:
\begin{enumerate}
    \item Monolingual setup: When training and development data is available for the same treebank/language, i.e. the goal is to train one model per treebank. It was allowed to use extra resources or tools that could boost performance, even from different languages.
    \item Cross-lingual, zero-shot setup: It is assumed that there is no gold training data in the language of the target treebank. The organizers specified that it is possible to use treebanks in other languages, translation tools, and any other resources that do not include sentiment annotations in the target language.
\end{enumerate}

\paragraph{Metrics} Each subtask is evaluated independently, and the ranking metric was \emph{sentiment graph F1} \cite{barnes-etal-2021-structured}, where true positives are exact matches at the graph level, weighting the overlap between the predicted and gold spans for each element, and averaged across all three spans. To compute precision, it
weights the number of correctly predicted tokens divided by the total number of predicted tokens, while for recall it weights the number of correctly predicted tokens divided by the total number of gold tokens. Also, as mentioned earlier, it is possible to have tuples with empty holders and targets.

\section{Our model}\label{section-model}

We rely on the \citet{dozat-manning-2018-simpler} parser, a widely used state-of-the-art model both for syntactic and semantic dependency parsing. Inspired in previous graph-based parsers \cite{mcdonald-etal-2005-non,kiperwasser-goldberg-2016-simple}, the parser first computes contextualized representations for each word using bidirectional LSTMs \citep[biLSTMs;][]{hochreiter1997long,schuster1997bidirectional}. After that, the model computes a \emph{head} and a \emph{dependent} representation for each term, to establish through a bi-affine attention whether an edge exists between each pair of tokens, and if so, what is the semantic relationship between them.
In particular, in this paper we follow the implementation used in the \texttt{supar}\footnote{\url{https://github.com/yzhangcs/parser}} package, as it has been widely adopted by the community and it is available for other flavors of parsing as well, such as constituent or dependency parsing. We preferred this implementation over the graph-based baseline provided in the SemEval repository, since early experiments showed a superior performance, and it also offered a simpler integration of large language models. We left the parser hyperparameters, except the learning rate, at their default value.

\paragraph{Pre-trained language models} For each language we looked for avaliable monolingual and multilingual pre-trained LMs at \url{https://huggingface.co/}. Specifically, for each language, we included:

\begin{itemize}
    \item \texttt{Basque}: berteus-base-cased, RoBasquERTa.
    \item \texttt{Catalan}: julibert, roberta-base-ca, calbert-base-uncased.
    \item \texttt{English}: bert-base-cased, bert-base-uncased, bert-large-cased, bert-large-uncased, roberta-base, roberta-large, albert-base-v2, albert-large-v2, xlnet-base-cased, xlnet-large-cased, electra-base-discriminator, electra-large-discriminator, electra-base-generator, electra-large-generator.
    \item \texttt{Norwegian}: norbert, nb-bert-base, nb-bert-large, electra-base-norwegian-uncased-discriminator.
    \item \texttt{Spanish}: bio-bert-base-spanish-wwm-uncased, bert-base-spanish-wwm-cased, roberta-base-bne, roberta-large-bne, selectra\_medium, zeroshot\_selectra\_medium.
\end{itemize}

\noindent With respect to the cross-lingual LMs, we considered: xlm-roberta-base and xlm-roberta-large. 

\subsection{Monolingual models}

For this task, we use: (i) pre-trained language models, (ii) \texttt{supar}, and (iii) the official training  and development files to build our models. Also note that we train end-to-end models, using words as the only input (later tokenized into subword pieces by the language models), but ignoring the part-of-speech tags and syntactic information provided in the sentiment treebanks. We did not use part-of-speech tags (or other morphosyntactic annotations) since these are not used in \texttt{supar} together with BERT encoders, and using them would require to adapt the code, which was exactly what we tried to avoid in this work.

\paragraph{Training procedure} We fine-tuned parsing models considering for each treebank the proposed LMs, and combining them with \texttt{supar}. Since training is time-consuming, many model configurations are proposed, and the performance of \texttt{supar} is stable independently of the seed, we decided to train a single model per LM. Specifically, all models have been trained with the default seed used by \texttt{supar}, which is 1. The only parameter that was modified was the learning rate (\textit{l\textsubscript{r}}), as we observed that for some models (specially the larger language models) the fine-tuning process did not converge. We started with \begin{math}5\cdot 10^{-5}\end{math}, and did a small grid search down to $1\cdot 10^{-6}$, where if a model still did not converge it was discarded.\footnote{For both monolingual and cross-lingual subtasks, all the selected models used a \textit{l\textsubscript{r}} of \begin{math}5\cdot 10^{-5}\end{math}. The only exception is Norwegian in the monolingual subtask, for which we used \begin{math}5\cdot 10^{-6}\end{math}.}
Additionally, to train the parsing models, we considered three strategies:
\begin{enumerate}
    \item \label{mono-approach-1} Single monolingual training and development files: We train each model on a single treebank and validate its performance in the corresponding dev set, i.e., the standard monolingual training and development methodology. 
\end{enumerate}

For the best model obtained for each treebank according to strategy \ref{mono-approach-1}, we explored a couple of harmonized training strategies (harmonized in the sense that different treebanks follow the same annotation guidelines):

\begin{enumerate}
\setcounter{enumi}{1}
    \item \label{mono-approach-2} Merged training and development files from different treebanks: We considered to merge all the available training files and all the available development files, treating them as a single dataset. Thus, we trained a single model that could predict all test files, but with the disadvantage that model selection is based on multilingual performance, which could hurt the performance in this setup.
    \item \label{mono-approach-3} Merged training files, single development file: Similar to \ref{mono-approach-2}, but merging only the training files. For the development phase, we proceeded as in \ref{mono-approach-1} and used each dataset's dev file for model selection. The idea was to have training data that can benefit from multilingual information, but that still considers a monolingual file for a language-dependent model selection, i.e., given $n$ treebanks, we still need to train $n$ models, one per treebank.
\end{enumerate}

We detail the experimental results for the training/development phase in \S \ref{section-results}.

\subsection{Cross-lingual (zero-shot) models}

In this setup, we rely on two main components: (i) available translation systems to perform word-level translations from source language to target language treebanks, and (ii) both monolingual and cross-lingual language models. Our goal with (i) is to obtain noisy, unlikely-grammatical data, but that still can provide sentiment annotations for a given target language, exploring the viability of this approach. Regarding the learning rate, we used $5\cdot 10^{-5}$ in all cases.

\paragraph{Auxiliary translation models} From the CoNLLU converted files\footnote{\url{https://github.com/MinionAttack/corpus-translator}}, we translated the sentences at the word level using the Helsinki-NLP translation models\footnote{\url{https://huggingface.co/Helsinki-NLP}} \cite{tiedemann-thottingal-2020-opus} available at \texttt{huggingface}. Table \ref{tab:crosslingual-translation-languages} lists the language pairs for which we could obtain translated versions for the cross-lingual setup.

\begin{table}[thbp!]
    \centering
    \scriptsize
    \begin{tabular}{l|cccc}
        \textbf{Dataset} & \textbf{Language} & \textbf{Basque} & \textbf{Catalan} & \textbf{Spanish} \\
        \hline
        NoReC\_fine & Norwegian &  &  & \checkmark \\ 
        MultiBooked & Basque &  &  & \checkmark \\
        MultiBooked & Catalan &  &  & \checkmark \\ 
        OpeNER & Spanish & \checkmark & \checkmark &  \\
        OpeNER & English & \checkmark & \checkmark & \checkmark \\ 
        MPQA & English & \checkmark & \checkmark & \checkmark \\
        Darmstadt\_unis & English & \checkmark & \checkmark & \checkmark \\
        \hline
    \end{tabular}
    \caption{Treebanks and the languages to which they were translated for the cross-lingual experiments.}
    \label{tab:crosslingual-translation-languages}
\end{table}

Then, to train the models we proceeded similarly to strategy \ref{mono-approach-2} used in the monolingual setup: we combined the translated training and validation files coming from treebanks in other languages, and used the micro-averaged F1-score on the translated development set for model selection.

\paragraph{Post-evaluation (and better) baseline} After the deadline to submit proposals, we also tested a baseline consisting on training, using an XLM-RoBERTa LM as the base component, a cross-lingual model that uses all the English datasets (without any kind of translation) as the source data. We discuss these results as well in \S \ref{section-crosslingual-task}.

\section{Results}\label{section-results}

Here, we detail and discuss the results that we got for both subtasks (see \S \ref{section-monolingual-task} and \ref{section-crosslingual-task}) on: (i) the official development sets, and (ii) the official test sets of the shared tasks.

\subsection{Monolingual setup}
\label{section-monolingual-task}

Tables \ref{tab:monolingual-dev-scores-english-before} and \ref{tab:monolingual-dev-scores-non-english-before} show the results for the development phase on the English and non-English datasets, respectively, including different LMs and training setups.

\begin{table}[hbtp!]
    \centering
    \scriptsize
    \begin{tabular}{c|c|c|c}
        \textbf{Corpus} & \textbf{Model} & \textbf{Strategy} & \textbf{F1} \\
        \hline
        OpeNER\_en & \makecell{xlm-roberta-large\\electra-base-discriminator\\xlm-roberta-large\\xlm-roberta-base\\roberta-large\\xlnet-base-cased\\xlm-roberta-base\\xlm-roberta-base\\roberta-base\\electra-large-discriminator\\bert-large-uncased\\electra-large-generator\\xlm-roberta-large\\bert-base-uncased\\bert-large-cased\\xlnet-large-cased\\bert-base-cased\\electra-base-generator\\albert-base-v2\\albert-large-v2} & \makecell{3\\1\\1\\2\\1\\1\\3\\1\\1\\1\\1\\1\\2\\1\\1\\1\\1\\1\\1\\1} & \makecell{0.714\\0.710\\0.707\\0.686\\0.683\\0.681\\0.679\\0.673\\0.663\\0.662\\0.660\\0.652\\0.643\\0.640\\0.640\\0.639\\0.612\\0.612\\0.590\\0.297} \\ 
        \hline
        MPQA & \makecell{roberta-base\\roberta-large\\electra-base-discriminator\\xlm-roberta-large\\xlnet-base-cased\\xlm-roberta-large\\bert-base-cased\\xlm-roberta-base\\electra-large-generator\\xlm-roberta-large\\bert-large-uncased\\bert-base-uncased\\xlm-roberta-base\\xlm-roberta-base\\bert-large-cased\\electra-base-generator\\albert-base-v2\\xlnet-large-cased} & \makecell{1\\1\\1\\1\\1\\3\\1\\2\\1\\2\\1\\1\\3\\1\\1\\1\\1\\1} & \makecell{0.374\\0.365\\0.351\\0.346\\0.338\\0.327\\0.306\\0.303\\0.301\\0.298\\0.297\\0.294\\0.285\\0.277\\0.269\\0.253\\0.236\\0.209} \\
        \hline
        Darmstadt\_unis & \makecell{xlm-roberta-large\\xlm-roberta-large\\electra-base-discriminator\\xlm-roberta-base\\xlm-roberta-large\\roberta-base\\xlm-roberta-base\\xlnet-base-cased\\xlnet-large-cased\\roberta-large\\electra-large-generator\\electra-large-discriminator\\xlm-roberta-base\\bert-large-uncased\\bert-base-uncased\\bert-large-cased\\electra-base-generator\\albert-base-v2\\bert-base-cased} & \makecell{3\\1\\1\\3\\2\\1\\1\\1\\1\\1\\1\\1\\2\\1\\1\\1\\1\\1\\1} & \makecell{0.329\\0.309\\0.306\\0.306\\0.301\\0.301\\0.276\\0.276\\0.269\\0.268\\0.267\\0.264\\0.262\\0.257\\0.251\\0.237\\0.229\\0.217\\0.212} \\
        \hline
    \end{tabular}
    \caption{Scores on the development set for the English treebanks and the monolingual setup. Models trained on the training data before its updated version.}
    \label{tab:monolingual-dev-scores-english-before}
\end{table}

\begin{table}[hbpt!]
    \centering
    \tiny
    \begin{tabular}{c|c|c|c}
        \textbf{Corpus} & \textbf{Model} & \textbf{Strategy} & \textbf{F1} \\
        \hline
        NoReC\_fine & \makecell{nb-bert-large\\nb-bert-base\\xlm-roberta-large\\xlm-roberta-large\\xlm-roberta-large\\xlm-roberta-base\\xlm-roberta-base\\xlm-roberta-base\\electra-base-norwegian\\-uncased-discriminator\\norbert} & \makecell{1\\1\\1\\3\\2\\3\\2\\1\\1\\\\1} & \makecell{0.479\\0.459\\0.450\\0.439\\0.427\\0.414\\0.411\\0.401\\0.382\\\\0.298} \\
        \hline
        MultiBooked\_eu & \makecell{xlm-roberta-large\\xlm-roberta-large\\xlm-roberta-base\\berteus-base-cased\\xlm-roberta-base\\xlm-roberta-base\\xlm-roberta-large\\RoBasquERTa} & \makecell{3\\2\\3\\1\\2\\1\\1\\1} & \makecell{0.662\\0.623\\0.613\\0.602\\0.597\\0.571\\0.569\\0.496} \\
        \hline
        MultiBooked\_ca & \makecell{xlm-roberta-base\\xlm-roberta-large\\xlm-roberta-large\\xlm-roberta-large\\xlm-roberta-base\\roberta-base-ca\\xlm-roberta-base\\julibert\\calbert-base-uncased} & \makecell{1\\1\\2\\3\\2\\1\\3\\1\\1} & \makecell{0.694\\0.683\\0.679\\0.679\\0.674\\0.672\\0.653\\0.590\\0.579} \\ 
        \hline
        OpeNER\_es & \makecell{xlm-roberta-large\\xlm-roberta-base\\xlm-roberta-base\\xlm-roberta-large\\xlm-roberta-base\\xlm-roberta-large\\bert-base-spanish-wwm-cased\\selectra\_medium\\roberta-base-bne\\zeroshot\_selectra\_medium\\roberta-large-bne\\bio-bert-base-spanish-wwm-uncased} & \makecell{3\\2\\3\\2\\1\\1\\1\\1\\1\\1\\1\\1} & \makecell{0.666\\0.662\\0.657\\0.639\\0.635\\0.635\\0.630\\0.622\\0.616\\0.610\\0.605\\0,457} \\
        \hline
    \end{tabular}
    \caption{Scores on the development set for the non-English treebanks and the monolingual setup. Models trained on the training data before its updated version.}
    \label{tab:monolingual-dev-scores-non-english-before}
\end{table}

With respect to the results on the English treebanks, an interesting trend is that despite being the monolingual setup, using cross-lingual language models, and in particular XLM-RoBERTa, performed surprisingly well. Combined with the training strategy \ref{mono-approach-3} (merged training sets, single development set), such models obtained the best results for 2 out of 3 English corpora (OpeNER and Darmstadt), while they still ranked well in the other dataset (MPQA). Across monolingual LMs, we also observe trends: electra-base-discriminator and (both base and large) RoBERTa models obtain overall the best results. On the other hand, we did not obtain equally robust results with ALBERT, and to a lesser extent, with BERT architectures. This is not totally surprising, since among the tested LMs, BERT is among the oldest ones, and ALBERT is a lite BERT, so some computational power is lost and it is understandable that this translates into some performance loss too, compared to larger LMs.

With respect to the experiments on the non-English datasets, we observe certain similarities, although the number of available models is much smaller than in the English cases. Again, XLM-RoBERTa overall obtains the best results. The only exception is the Norwegian dataset, where we obtained the best results with a BERT architecture. 

Yet, a more thoughtful discussion would be needed to determine if some architectures truly behave better than others. Note that all these LMs are usually pre-trained using different and heterogeneous text sources, and specially for the less-resourced languages, some constraints are usually imposed during training. For instance, it is hard to conclude that berteus-base-cased (BERT) \cite{agerri-etal-2020-give} is worse than XLM-RoBERTa \cite{conneau-etal-2020-unsupervised}, since the amount of resources to train the former was more constrained.

Finally, a few days before the submission deadline, the training files of some treebanks were slightly updated by the organizers, due to minor bugs in the segmentation process that corrupted some sentences.  As we did not have time to rerun all models and update the results, we chose to re-train only the model that obtained the best performance on the previous version of the treebanks. Therefore, all the outputs submitted for the test sets correspond to models trained on the updated, uncorrupted files. In Table \ref{tab:monolingual-dev-scores-after} we compare the performance of the models trained on the updated and deprecated versions of the training files. Overall, we observed relatively small, but non-negligible differences, usually obtaining a better performance with the updated version of the treebank.

\begin{table}[thbp!]
    \centering
    \scriptsize
    \begin{tabular}{c|c|c|c|c}
        \textbf{Corpus} & \textbf{Model} & \textbf{Strategy} & \textbf{Old F1} & \textbf{New F1} \\
        \hline
        NoReC\_fine & nb-bert-large & 1 & 0.479 & 0.492 \\
        MultiBooked\_eu & xlm-roberta-large & 3 & 0.662 & 0.648 \\
        MultiBooked\_ca & xlm-roberta-base & 1 & 0.694 & 0.699 \\
        OpeNER\_es & xlm-roberta-large & 3 & 0.666 & 0.709 \\
        OpeNER\_en & xlm-roberta-large & 3 & 0.714 & 0.716 \\
        MPQA & roberta-base & 1 & 0.374 & 0.374 \\
        Darmstadt\_unis & xlm-roberta-large & 3 & 0.329 & 0.357 \\
        \hline
    \end{tabular}
    \caption{Scores on the development set for the models trained on the corrupted and uncorrupted versions of the training files, on the monolingual setup. For each treebank, we only did the comparison for the best performing model, based on the performance on the corrupted version.}
    \label{tab:monolingual-dev-scores-after}
\end{table}

\paragraph{Official results on the test sets} In Table \ref{tab:model-test-scores-monolingual} we show the performance on the test sets of our submitted models, i.e. those that achieved the highest score in the corresponding development phase. The performance is stable across different test sets, obtaining slightly better results for Iberian languages. For a detailed comparison against the rest of participants, we refer the users to Appendix \ref{tab:monolingual-leaderboard} and the official shared task paper \cite{barnes-etal-2022-semeval}.

\begin{table}[t]
    \centering
    \scriptsize
    \begin{tabular}{c|c|c|c}
        \textbf{Dataset} & \textbf{Model} & \textbf{Strategy} & \textbf{Score} \\
    \hline
        NoReC\_fine & nb-bert-large & 1 & 0.462\textsubscript{(10)} \\
        MultiBooked\_eu & xlm-roberta-large & 2 & 0.680\textsubscript{(7)} \\
        MultiBooked\_ca & xlm-roberta-base & 1 & 0.653\textsubscript{(8)} \\
        OpeNER\_es & xlm-roberta-large & 3 & 0.692\textsubscript{(6)} \\
        OpeNER\_en & xlm-roberta-large & 3 & 0.698\textsubscript{(9)} \\
        MPQA & roberta-base & 1 & 0.349\textsubscript{(10)} \\
        Darmstadt\_unis & xlm-roberta-large & 3 & 0.414\textsubscript{(8)} \\
        \hline
    \end{tabular}
    \caption{Scores of our models, for the monolingual subtask, on each test set. Our ranking on the shared task for each test set is indicated as a subscript.}
    \label{tab:model-test-scores-monolingual}
\end{table}

The datasets of the shared task belong to different domains: OpeNER and MultiBooked deal with hotel reviews, NoReC with professional reviews in multiple domains, Darmstadt\_unis (the dataset for which we obtain the second lowest scores) contains English online university reviews, and MPQA (the dataset for which we obtain the lowest scores) is about news articles annotated with opinions and other private states
(i.e., beliefs, emotions, sentiments, speculations, \dots).
For the two lowest-scoring datasets, they have in common that they mostly contain single-opinion sentences, whereas the other datasets tend to have more variety in the number of opinions and their distribution. For instance, $\sim$85\% and $\sim$74\%  of the training sentences of the Darmstadt\_unis and MPQA datasets have only one opinion, while the next most `single-opinion' dataset is multibooked\_eu with only $\sim$53\% of the sentences. However, we need to perform more detailed analysis as future work to extract more robust conclusions.

\subsection{Cross-lingual setup}\label{section-crosslingual-task}

Table \ref{tab:crosslingual-dev-scores-before} shows the results for the development phase for the three target languages and their datasets. Again, XLM-RoBERTa models obtain overall the best performance, although in this case it is less surprising since cross-lingual LMs are expected to suit well this kind of challenges.
Similar to the case of the monolingual setup, we decided to retrain the best-performing model with the updated versions of the training files. In Table \ref{tab:crosslingual-dev-scores-after}, we show the comparison between the corrupted and uncorrupted versions of the datasets, which contrarily to the monolingual setup, often turned out into worse performing models.

\begin{table}[t]
    \centering
    \small
    \begin{tabular}{c|c|c}
        \textbf{Corpus} & \textbf{Model} & \textbf{F1} \\
        \hline
        Basque & \makecell{xlm-roberta-base\\berteus-base-cased\\RoBasquERTa} & \makecell{0.434\\0.416\\0.323} \\
        \hline
        Catalan & \makecell{roberta-base-ca\\xlm-roberta-base\\julibert\\calbert-base-uncased} & \makecell{0.564\\0.519\\0.486\\0.385} \\
        \hline
        Spanish & \makecell{xlm-roberta-base\\xlm-roberta-large\\bert-base-spanish-wwm-cased\\zeroshot\_selectra\_medium\\selectra\_medium\\roberta-base-bne\\roberta-large-bne\\bio-bert-base-spanish-wwm-uncased} & \makecell{0.605\\0.593\\0.583\\0.555\\0.536\\0.515\\0.438\\0.386} \\ 
        \hline
    \end{tabular}
    \caption{Scores on the development set for the \emph{translated} English treebanks and the cross-lingual setup. Models trained on the training data before its updated version.}
    \label{tab:crosslingual-dev-scores-before}
\end{table}

\begin{table}[t]
    \centering
    \small
    \begin{tabular}{l|c|c|c}
        \textbf{Language} & \textbf{Model} & \textbf{Old F1} & \textbf{New F1} \\
        \hline
        Basque & \makecell{berteus-base-cased\\xlm-roberta-base} & \makecell{0.416\\0.434} & \makecell{0.424\\0.416} \\
        \hline
        Catalan & roberta-base-ca & 0.564 & 0.572 \\
        \hline
        Spanish & \makecell{xlm-roberta-large\\xlm-roberta-base} & \makecell{0.593\\0.605} & \makecell{0.570\\0.569} \\
        \hline
    \end{tabular}
    \caption{Scores on the development set for the models trained on the corrupted and uncorrupted versions of the \emph{translated} training files, on the cross-lingual setup. For each treebank, we only did the comparison for the best performing model, based on the performance on the corrupted version.}
    \label{tab:crosslingual-dev-scores-after}
\end{table}

Finally, Table \ref{tab:crosslingual-dev-scores-from-english} shows the scores for the post-evaluation baseline (model trained on the English datasets with XLM-RoBERTa) on the dev set.
Very interestingly, the results show that this baseline outperformed our word-level translation approaches. We need more analysis to understand why this happens, but we hypothesize that the larger amount of English texts XLM-RoBERTa was pre-trained on could be playing an important role.

\begin{table}[thbp!]
    \centering
    \small
    \begin{tabular}{l|c|c}
        \textbf{Language} & \textbf{Model} & \textbf{Score} \\
        \hline
         Basque & \makecell{xlm-roberta-base\\xlm-roberta-large} & \makecell{0.678\\0.677} \\
        \hline
         Catalan & \makecell{xlm-roberta-base\\xlm-roberta-large} & \makecell{0.598\\0.625} \\
        \hline
         Spanish & \makecell{xlm-roberta-base\\xlm-roberta-large} & \makecell{0.663\\0.638} \\
        \hline
    \end{tabular}
    \caption{Scores on the development set of the trained English models (trained on MPQA, OpeNER\_en and Darmstadt\_unis corpora, \emph{without} word-level translation) for the cross-lingual subtask.}
    \label{tab:crosslingual-dev-scores-from-english}
\end{table}

\paragraph{Official results on the test sets} Finally, in Table \ref{tab:model-test-scores-crosslingual} we show our results on test sets of the cross-lingual, zero-shot setup, for which we obtain again stable results. Appendix \ref{tab:crosslingual-leaderboard} contains the results for all participants.

\begin{table}[thbp!]
    \centering
    \small
    \begin{tabular}{c|c|c}
        \textbf{Language} & \textbf{Model} & \textbf{Score} \\
        \hline
        \multicolumn{3}{l}{\textbf{\color{gray}\scriptsize Target language: Basque}} \\
        \hline
        Basque & berteus-base-cased & 0.509\textsubscript{(8)} \\
        \hline
        Catalan & roberta-base-ca & 0.554\textsubscript{(8)} \\
        \hline
        Spanish & xlm-roberta-large & 0.570\textsubscript{(7)} \\
        \hline
        \multicolumn{3}{l}{\textbf{\color{gray}\scriptsize Combined English corpora without word-level translation}} \\
        \hline
        Basque & \makecell{xlm-roberta-base\\xlm-roberta-large} & \makecell{0.649\textsubscript{(2)*}\\0.641\textsubscript{(2)*}} \\
        \hline
        Catalan & \makecell{xlm-roberta-base\\xlm-roberta-large} & \makecell{0.647\textsubscript{(2)*}\\0.655\textsubscript{(2)*}} \\
        \hline
        Spanish & \makecell{xlm-roberta-base\\xlm-roberta-large} & \makecell{0.670\textsubscript{(1)*}\\0.638\textsubscript{(2)*}} \\
        \hline
    \end{tabular}
    \caption{Scores of our models, for the cross-lingual subtask, on each test set. Our ranking on the shared task for each test set is indicated as a subscript. * indicates the ranking that we would obtain in the shared task using the post-evaluation baseline models.}
    \label{tab:model-test-scores-crosslingual}
\end{table}

\section{Conclusion}

This paper describes our participation at the Sem\-Eval Shared Task 10 on structured sentiment analysis. We participated both in the monolingual and cross-lingual (zero-shot) setups. We applied a simple, but effective approach, relying on off-the-shelf tools, traditionally used for other purposes, and used them to predict sentiment graphs instead. More particularly, for the monolingual setup, we linked pre-trained language models with bi-affine graph parsing and training over single and multiple treebanks. In the zero-shot setup, we followed a similar approach, but relied on publicly available translation models to obtain training data, by applying a word-level translation of treebanks, to then train models similarly to the monolingual setup.

\section*{Acknowledgements}

This work is supported by a 2020 Leonardo Grant for Researchers and Cultural Creators from the FBBVA,\footnote{FBBVA accepts no responsibility for the opinions, statements and contents included in the project and/or the results thereof, which are entirely the responsibility of the authors.} as well as by the European Research Council (ERC), under the European Union’s Horizon 2020 research and innovation programme (FASTPARSE, grant agreement No 714150). The work is also supported by ERDF/MICINN-AEI (SCANNER-UDC, PID2020-113230RB-C21), by Xunta de Galicia (ED431C 2020/11), and by Centro de Investigación de Galicia ‘‘CITIC’’ which is funded by Xunta de Galicia, Spain and the European Union (ERDF - Galicia 2014–2020 Program), by grant ED431G 2019/01.

\bibliography{anthology,custom}
\bibliographystyle{acl_natbib}

\appendix

\section{Shared Task Leaderboard}
\label{sec:shared-task-tLeaderboard}

\begin{table*}[thbp!]
    \centering
    \tiny
    \tabcolsep=0.11cm
    \begin{tabular}{c c c c c c c c c c c}
        \textbf{User} & \textbf{Team} & \textbf{NoReC\_fine} & \textbf{MultiBooked\_ca} & \textbf{MultiBooked\_eu} & \textbf{OpeNER\_en} & \textbf{OpeNER\_es} & \textbf{MPQA} & \textbf{Darmstadt\_unis} & \textbf{Average} \\ \rowcolor[gray]{.95}
        \hline
        zhixiaobao &  & 0.529 (2) & 0.728 (1) & 0.739 (1) & 0.760 (2) & 0.722 (4) & 0.447 (1) & 0.494 (1) & 0.631 (1) \\
        Cong666 &  & 0.524 (3) & 0.728 (1) & 0.739 (1) & 0.763 (1) & 0.742 (1) & 0.416 (2) & 0.485 (2) & 0.628 (2) \\ \rowcolor[gray]{.95}
        gmorio & Hitachi & 0.533 (1) & 0.709 (3) & 0.715 (3) & 0.756 (3) & 0.732 (3) & 0.402 (3) & 0.463 (3) & 0.616 (3) \\
        colorful &  & 0.497 (5) & 0.678 (6) & 0.723 (2) & 0.745 (4) & 0.735 (2) & 0.375 (5) & 0.380 (12) & 0.590 (4) \\ \rowcolor[gray]{.95}
        whu\_stone & sixsixsix & 0.483 (9) & 0.711 (2) & 0.681 (6) & 0.727 (6) & 0.686 (7) & 0.379 (4) & 0.373 (13) & 0.577 (5) \\
        KE\_AI &  & 0.483 (9) & 0.711 (2) & 0.681 (6) & 0.727 (6) & 0.686 (7) & 0.364 (7) & 0.373 (13) & 0.575 (6) \\ \rowcolor[gray]{.95}
        Fadi & SeqL & 0.488 (7) & 0.699 (4) & 0.701 (4) & 0.730 (5) & 0.700 (5) & 0.245 (20) & 0.394 (11) & 0.565 (7) \\
        lys\_acoruna & LyS\_ACoruña & 0.462 (10) & 0.653 (8) & 0.680 (7) & 0.698 (9) & 0.692 (6) & 0.349 (10) & 0.414 (7) & 0.564 (8) \\ \rowcolor[gray]{.95}
        QiZhang & ECNU\_ICA & 0.496 (6) & 0.684 (5) & 0.686 (5) & 0.676 (10) & 0.623 (11) & 0.351 (8) & 0.409 (8) & 0.561 (9) \\
        luxinyu & ohhhmygosh & 0.487 (8) & 0.658 (7) & 0.651 (9) & 0.710 (7) & 0.669 (8) & 0.269 (19) & 0.416 (6) & 0.551 (10) \\ \rowcolor[gray]{.95}
        rafalposwiata & OPI & 0.459 (11) & 0.650 (9) & 0.653 (8) & 0.670 (11) & 0.663 (9) & 0.326 (13) & 0.395 (10) & 0.545 (11) \\
        evanyfyang &  & 0.213 (22) & 0.635 (11) & 0.639 (10) & 0.703 (8) & 0.642 (10) & 0.350 (9) & 0.449 (4) & 0.519 (12) \\ \rowcolor[gray]{.95}
        robvanderg &  & 0.366 (13) & 0.648 (10) & 0.605 (11) & 0.632 (14) & 0.614 (13) & 0.296 (15) & 0.344 (14) & 0.501 (13) \\
        psarangi & AMEX AI Labs & 0.343 (15) & 0.634 (12) & 0.559 (12) & 0.634 (13) & 0.595 (14) & 0.283 (17) & 0.320 (17) & 0.481 (14) \\ \rowcolor[gray]{.95}
        chx.dou & abondoned & 0.395 (12) & 0.583 (13) & 0.506 (13) & 0.626 (15) & 0.622 (12) & 0.309 (14) & 0.280 (19) & 0.474 (15) \\
        zaizhep & MMAI & 0.329 (16) & 0.525 (14) & 0.478 (17) & 0.623 (16) & 0.539 (16) & 0.367 (6) & 0.342 (15) & 0.458 (16) \\ \rowcolor[gray]{.95}
        janpf &  & 0.280 (19) & 0.517 (15) & 0.439 (19) & 0.651 (12) & 0.504 (17) & 0.338 (11) & 0.417 (5) & 0.449 (17) \\
        etms.kgp & ETMS@IITKGP & 0.351 (14) & 0.508 (16) & 0.438 (20) & 0.626 (15) & 0.544 (15) & 0.327 (12) & 0.330 (16) & 0.446 (18) \\ \rowcolor[gray]{.95}
        jylong &  & 0.323 (18) & 0.474 (19) & 0.504 (14) & 0.476 (17) & 0.375 (21) & 0.274 (18) & 0.223 (21) & 0.379 (19) \\
        ouzh &  & 0.323 (18) & 0.474 (19) & 0.504 (14) & 0.476 (17) & 0.375 (21) & 0.274 (18) & 0.223 (21) & 0.378 (20) \\ \rowcolor[gray]{.95}
        SPDB\_Innovation\_Lab & Innovation Lab & 0.325 (17) & 0.469 (20) & 0.486 (16) & 0.471 (18) & 0.362 (22) & 0.289 (16) & 0.202 (22) & 0.372 (21) \\
        lucasrafaelc &  & 0.251 (21) & 0.505 (17) & 0.467 (18) & 0.431 (19) & 0.399 (19) & 0.232 (21) & 0.230 (20) & 0.359 (22) \\ \rowcolor[gray]{.95}
        foodchup &  & 0.265 (20) & 0.493 (18) & 0.491 (15) & 0.415 (20) & 0.480 (18) & 0.149 (22) & 0.139 (24) & 0.347 (23) \\
        jzh1qaz &  & 0.186 (25) & 0.431 (21) & 0.385 (21) & 0.381 (21) & 0.393 (20) & 0.094 (23) & 0.092 (26) & 0.280 (24) \\ \rowcolor[gray]{.95}
        hades\_d & Mirs & 0.504 (4) & 0.678 (6) & 0.000 (25) & 0.000 (25) & 0.000 (26) & 0.375 (5) & 0.400 (9) & 0.280 (24) \\
        huyenbui117 &  & 0.194 (23) & 0.341 (22) & 0.374 (22) & 0.316 (23) & 0.245 (25) & 0.009 (26) & 0.053 (27) & 0.219 (25) \\ \rowcolor[gray]{.95}
        karun842002 & SSN\_MLRG1 & 0.191 (24) & 0.323 (23) & 0.331 (23) & 0.306 (24) & 0.257 (24) & 0.015 (25) & 0.104 (25) & 0.218 (26) \\
        gerarld & nlp2077 & 0.000 (26) & 0.269 (24) & 0.303 (24) & 0.354 (22) & 0.321 (23) & 0.019 (24) & 0.180 (23) & 0.207 (27) \\ \rowcolor[gray]{.95}
        michael\_wzhu91 & kobe4ever & 0.000 (26) & 0.000 (25) & 0.000 (25) & 0.000 (25) & 0.000 (26) & 0.000 (27) & 0.306 (18) & 0.044 (28) \\
        normalkim &  & 0.000 (26) & 0.000 (25) & 0.000 (25) & 0.000 (25) & 0.000 (26) & 0.000 (27) & 0.000 (28) & 0.000 (29) \\ \rowcolor[gray]{.95}
        UniParma & UniParma & 0.000 (26) & 0.000 (25) & 0.000 (25) & 0.000 (25) & 0.000 (26) & 0.000 (27) & 0.000 (28) & 0.000 (29) \\
        whu\_venti &  & 0.000 (26) & 0.000 (25) & 0.000 (25) & 0.000 (25) & 0.000 (26) & 0.000 (27) & 0.000 (28) & 0.000 (29) \\
    \end{tabular}
    \caption{Leaderboard of all participants in the monolingual task}
    \label{tab:monolingual-leaderboard}
\end{table*}

\begin{table*}[thbp!]
    \centering
    \small
    \tabcolsep=0.11cm
    \begin{tabular}{c c c c c c c c c c c}
        \textbf{User} & \textbf{Team} & \textbf{OpeNER\_es} & \textbf{MultiBooked\_ca} & \textbf{MultiBooked\_eu} & \textbf{Average} \\ \rowcolor[gray]{.95}
        \hline
        Cong666 &  & 0.644 (1) & 0.643 (1) & 0.632 (1) & 0.640 (1) \\
        luxinyu & ohhhmygosh & 0.620 (3) & 0.605 (4) & 0.569 (2 &  0.598 (2) \\ \rowcolor[gray]{.95}
        gmorio & Hitachi & 0.628 (2) & 0.607 (3) & 0.527 (4) & 0.587 (3) \\
        whu\_stone & sixsixsix & 0.604 (5) & 0.596 (5) & 0.512 (7) & 0.571 (4) \\ \rowcolor[gray]{.95}
        QiZhang & ECNU\_ICA & 0.551 (10) & 0.615 (2) & 0.530 (3) & 0.566 (5) \\
        Fadi & SeqL & 0.589 (6) & 0.593 (6) & 0.516 (6) & 0.566 (5) \\ \rowcolor[gray]{.95}
        colorful &  & 0.620 (3) & 0.543 (11) & 0.527 (4) & 0.563 (6) \\
        hades\_d & Mirs & 0.617 (4) & 0.544 (10) & 0.522 (5) & 0.561 (7) \\ \rowcolor[gray]{.95}
        lys\_acoruna & LyS\_ACoruña & 0.570 (7) & 0.554 (8) & 0.509 (8) & 0.544 (8) \\
        rafalposwiata & OPI & 0.564 (8) & 0.586 (7) & 0.444 (12) & 0.531 (9) \\ \rowcolor[gray]{.95}
        KE\_AI &  & 0.561 (9) & 0.552 (9) & 0.463 (11) & 0.525 (10) \\
        etms.kgp & ETMS@IITKGP & 0.542 (11) & 0.506 (12) & 0.431 (13) & 0.493 (11) \\ \rowcolor[gray]{.95}
        jylong &  & 0.375 (12) & 0.474 (13) & 0.504 (9) & 0.451 (12) \\
        ouzh &  & 0.375 (12) & 0.474 (13) & 0.504 (9) & 0.451 (12) \\ \rowcolor[gray]{.95}
        SPDB\_Innovation\_Lab & SPDB Innovation Lab & 0.362 (13) & 0.469 (14) & 0.486 (10) & 0.439 (13) \\
        gerarld & nlp2077 & 0.321 (14) & 0.269 (15) & 0.303 (14) & 0.298 (14) \\ \rowcolor[gray]{.95}
        janpf &  & 0.315 (15) & 0.259 (16) & 0.243 (15) & 0.272 (15) \\
        chx.dou & abondoned & 0.013 (16) & 0.009 (17) & 0.004 (16) & 0.009 (16) \\ \rowcolor[gray]{.95}
        jzh1qaz &  & 0.000 (17) & 0.000 (18) & 0.000 (17) & 0.000 (17) \\
        zhixiaobao &  & 0.000 (17) & 0.000 (18) & 0.000 (17) & 0.000 (17) \\ \rowcolor[gray]{.95}
        psarangi & AMEX AI Labs & 0.000 (17) & 0.000 (18) & 0.000 (17) & 0.000 (17) \\
        normalkim &  & 0.000 (17) & 0.000 (18) & 0.000 (17) & 0.000 (17) \\ \rowcolor[gray]{.95}
        zaizhep & MMAI & 0.000 (17) & 0.000 (18) & 0.000 (17) & 0.000 (17) \\
        lucasrafaelc &  & 0.000 (17) & 0.000 (18) & 0.000 (17) & 0.000 (17) \\ \rowcolor[gray]{.95}
        evanyfyang &  & 0.000 (17) & 0.000 (18) & 0.000 (17) & 0.000 (17) \\
        robvanderg &  & 0.000 (17) & 0.000 (18) & 0.000 (17) & 0.000 (17) \\ \rowcolor[gray]{.95}
        michael\_wzhu91 & kobe4ever & 0.000 (17) & 0.000 (18) & 0.000 (17) & 0.000 (17) \\
        UniParma & UniParma & 0.000 (17) & 0.000 (18) & 0.000 (17) & 0.000 (17) \\ \rowcolor[gray]{.95}
        huyenbui117 &  & 0.000 (17) & 0.000 (18) & 0.000 (17) & 0.000 (17) \\
        karun842002 & SSN\_MLRG1 & 0.000 (17) & 0.000 (18) & 0.000 (17) & 0.000 (17) \\ \rowcolor[gray]{.95}
        whu\_venti &  & 0.000 (17) & 0.000 (18) & 0.000 (17) & 0.000 (17) \\
        foodchup &  & 0.000 (17) & 0.000 (18) & 0.000 (17) & 0.000 (17) \\
    \end{tabular}
    \caption{Leaderboard of all participants in the cross-lingual task}
    \label{tab:crosslingual-leaderboard}
\end{table*}

\end{document}